\documentclass{article}
\usepackage{arxiv}          

\usepackage[T1]{fontenc}
\usepackage[utf8]{inputenc}
\usepackage{graphicx}
\usepackage{booktabs}
\usepackage{amsmath,amssymb}
\usepackage{tikz}
\usetikzlibrary{positioning,arrows.meta}
\usepackage[numbers,sort&compress]{natbib}   
\usepackage[hidelinks]{hyperref}
\usepackage{url}
\usepackage{enumitem}

\usepackage[acronym, nogroupskip, nonumberlist, nopostdot]{glossaries}
\definecolor{myblue}{RGB}{0,0,255}

\newacronym{aadl}{AADL}{Architecture Analysis and Design Language}
\newacronym{ansi-c}{ANSI-C}{American National Standards Institute C}
\newacronym{api}{API}{Application Programming Interface}
\newacronym{adc}{ADC}{Analog-to-Digital Converter}
\newacronym{ast}{AST}{Abstract Syntax Tree}
\newacronym{bdd}{BDD}{Binary Decision Diagrams}
\newacronym{bmc}{BMC}{Bounded Model Checking}
\newacronym{cbmc}{CBMC}{Bounded Model Checking for ANSI-C Programs}
\newacronym{cegar}{CEGAR}{Counterexample-Guided Abstraction Refinement}
\newacronym{cern}{CERN}{Conseil Européen pour la Recherche Nucléaire}
\newacronym{cfg}{CFG}{Control Flow Graph}
\newacronym{chc}{CHC}{Constrained Horn Clause}
\newacronym{cli}{CLI}{Command-Line Interface}
\newacronym{cpu}{CPU}{Central Processing Unit}
\newacronym{ctl}{CTL}{Computation Tree Logic}
\newacronym{cuda}{CUDA}{Compute Unified Device Architecture}
\newacronym{cve}{CVE}{Common Vulnerability and Exposure}
\newacronym{dfs}{DFS}{Depth-First Search}
\newacronym{dsl}{DSL}{Domain-Specific Language}
\newacronym{epsrc}{EPSRC}{Engineering and Physical Sciences Research Council}
\newacronym{evm}{EVM}{Ethereum Virtual Machine}
\newacronym{esbmc}{ESBMC}{Efficient SMT-based Context-Bounded Model Checker}
\newacronym{fbd}{FBD}{Functional Block Diagram}
\newacronym{fpga}{FPGA}{Field-Programmable Gate Array}
\newacronym{gpio}{GPIO}{General-Purpose Input/Output}
\newacronym{hal}{HAL}{Hardware Abstraction Layer}
\newacronym{hil}{HIL}{Hardware-in-the-Loop}
\newacronym{ic3}{IC3}{Incremental Construction of Inductive Clauses for Indubitable Correctness}
\newacronym{ide}{IDE}{Integrated Development Environment}
\newacronym{iec}{IEC}{International Electrotechnical Commission}
\newacronym{ieee}{IEEE}{Institute of Electrical and Electronics Engineers}
\newacronym{ics}{ICS}{Industrial Control Systems}
\newacronym{il}{IL}{Instruction List}
\newacronym{iot}{IoT}{Internet of Things}
\newacronym{ir}{IR}{Intermediate Representation}
\newacronym{iso}{ISO}{International Organization for Standardization}
\newacronym{ld}{LD}{Ladder Diagram}
\newacronym{llb}{LLB}{Ladder Logic Bombs}
\newacronym{llm}{LLM}{Large Language Model}
\newacronym{ltl}{LTL}{Linear Temporal Logic}
\newacronym{matiec}{MATIEC}{IEC 61131-3 compiler}
\newacronym{mcu}{MCU}{Microcontroller Unit}
\newacronym{nasa}{NASA}{National Aeronautics and Space Administration}
\newacronym{pwm}{PWM}{Pulse-Width Modulation}
\newacronym{pdr}{PDR}{Property Directed Reachability}
\newacronym{pid}{PID}{Proportional-Integral-Derivative}
\newacronym{plc}{PLC}{Programmable Logic Controller}
\newacronym{pou}{POU}{Program Organization Unit}
\newacronym{por}{POR}{Partial Order Reduction}
\newacronym{ppu}{PPU}{Pick and Place Unit}
\newacronym{rtos}{RTOS}{Real-Time Operating System}
\newacronym{sat}{SAT}{Boolean Satisfiability}
\newacronym{scl}{SCL}{Structured Control Language}
\newacronym{sfc}{SFC}{Sequential Function Chart}
\newacronym{slr}{SLR}{Systematic Literature Review}
\newacronym{smt}{SMT}{Satisfiability Modulo Theories}
\newacronym{smtlib2}{SMT-LIB}{Satisfiability Modulo Theories Library}
\newacronym{smv}{SMV}{Symbolic Model Verifier}
\newacronym{ssa}{SSA}{Static Single Assignment}
\newacronym{st}{ST}{Structured Text}
\newacronym{stl}{STL}{Statement List}
\newacronym{svcomp}{SV-COMP}{Competition on Software Verification}
\newacronym{tacas}{TACAS}{Tools and Algorithms for the Construction and Analysis of Systems}
\newacronym{ufam}{UFAM}{Federal University of Amazonas}
\newacronym{ukri}{UKRI}{UK Research and Innovation}

\providecommand{\esbmcplc}{\textsc{\mbox{ESBMC-PLC}}}
\providecommand{\plcld}{\textsc{\mbox{PLC-LD}}}
\providecommand{\plcopen}{PLCopen}

\begin{document}

\title{A Benchmark Suite and Ground-Truth Methodology for Formal Verification of IEC~61131-3 Ladder Diagram Programs}

\renewcommand{\shorttitle}{A Benchmark Suite for Formal Verification of IEC~61131-3 LD}

\date{}

\author{
  Pierre Dantas\thanks{Corresponding author.} \\
  Department of Computer Science \\
  University of Manchester \\
  Manchester, UK \\
  \texttt{pierre.dantas@manchester.ac.uk} \\
  ORCID: 0000-0001-6390-9340
  \And
  Lucas C. Cordeiro \\
  Department of Computer Science \\
  University of Manchester \\
  Manchester, UK \\
  \texttt{lucas.cordeiro@manchester.ac.uk} \\
  ORCID: 0000-0002-6235-4272
  \And
  Waldir Junior \\
  Electrical Engineering \\
  Federal University of Amazonas (UFAM) \\
  Manaus, AM, Brazil \\
  \texttt{waldirjr@ufam.edu.br} \\
  ORCID: 0000-0003-3095-0042
}

\maketitle
\begin{abstract}
\glspl{plc} govern safety-critical physical infrastructure, and a growing set of tools now applies formal verification to their IEC~61131-3 programs. Yet the field lacks a standard benchmark suite, so tools are evaluated on private, incomparable program sets, and progress cannot be measured. Existing resources are either program corpora without formal properties or expected verdicts, or a single \gls{st} suite that omits the graphical \gls{ld} encoding used to exchange real programs. We present a benchmark suite of \textbf{50 programs over 83 variants}, spanning textual and graphical \gls{ld} and \gls{st} across ten industrial domains, each in standard PLCopen~XML or \gls{st} and paired with a formal property, a machine-checkable expected verdict, and, for every violation, a triggering witness, in an SV-COMP--compatible layout. Our central contribution is a \emph{ground-truth methodology}: every verdict is established by one of three recorded methods (construction, fault injection, or audited cross-tool consensus), a discipline motivated by a concrete episode in which the ``obvious'' safety property would have mislabelled every attack in two public logic-bomb corpora as safe, because the injected defect is a non-terminating loop that the obvious property cannot see. We establish reference verdicts with a from-source build of ESBMC~v8.4: all 25 graphical benchmarks run, and 43 of 45 accepted variants match the recorded verdict with no disagreements. On the finite-state fragment it can ingest (21 of the 50 benchmarks), an independent model checker (nuXmv), whose decision procedure is unrelated to ESBMC's, agrees on all 24 interlock variants and resolves two stateful benchmarks that ESBMC-PLC leaves unknown -- evidence both that the ground truth is tool-neutral and that the suite discriminates between tools. Porting the textual and structured-text benchmarks surfaces a \emph{format-and-semantics} fragmentation -- one tool's front-end admits a single serialization, and even the meaning of a standard timer is not agreed across tools -- which is itself the phenomenon the suite is designed to expose. The corpus, schema, validator, and recheck harness are released as open artifacts. 

\end{abstract}


\glsresetall

\section{Introduction}
\label{sec:intro}

\glspl{plc} run the logic of safety-critical physical infrastructure -- water treatment, power distribution, manufacturing, rail, and building automation -- where a control-logic defect can cause equipment damage, environmental release, or injury. The dominant programming standard for these devices, IEC~61131-3, and in particular its \gls{ld} language, is therefore a natural and important target for formal verification. A growing body of tools now attempts it: \esbmcplc{} and its successors, PLCverif at \gls{cern}, and model checkers such as nuXmv, among others.

Yet the field cannot easily measure its own progress. Unlike software verification for C, where the SV-COMP benchmark suite and competition turned a collection of isolated tools into a comparable, cumulatively improving community, \gls{plc} verification has \emph{no standard benchmark suite}. Tool papers evaluate on private or incomparable program sets, so claims of coverage, precision, and extensibility cannot be compared, and no shared target exists against which a new technique can demonstrate improvement. The resources that do exist are lacking in one of two ways. Program corpora -- such as the \plcld{} tank-control dataset or the \gls{ppu} clone-detection set -- ship code but no formal properties and no expected verdicts, so they are not verification assignments at all. And the one property-bearing suite we are aware of covers \gls{st} only, omitting graphical \gls{ld} -- the encoding in which real programs are exchanged via \plcopen{}~XML and which dominates the installed base. Building such a suite is not simply a matter of collecting programs: it requires attaching a formal property and a \emph{trustworthy} expected verdict to each program, and doing so in a format that any tool can run.

This paper presents that suite: \textbf{50 benchmarks over 83 program variants}, spanning textual and graphical \gls{ld} and \gls{st}, across ten industrial fields, each in standard \plcopen{}~XML or \gls{st} and paired with a formal property, a machine-checkable expected verdict, and, for every violation, a triggering witness. The suite is packaged in an SV-COMP--compatible layout with a schema validator and a one-command recheck harness, so a new tool is added through an adapter rather than by reformatting the corpus. So the artifact can directly seed a \gls{plc}/\gls{ics} category in SV-COMP.

The main challenge, and our main methodological contribution, is \emph{ground truth}. An expected verdict that is silently wrong is worse than no benchmark, because it rewards incorrect answers, and for programs of any realism, the correct verdict is not obvious. One episode from building the suite makes the point concretely. Two public logic-bomb corpora supply pairs of a ``legitimate'' program and a ``malicious'' sibling, an apparently ready source of safe/violation labels. The obvious safety property for these tank controllers is a valve interlock -- and asserting it would have been a mistake: reading the injected code reveals that the malicious variants leave the valve logic \emph{untouched} and instead insert a non-terminating loop, so the interlock holds in both and the ``obvious'' property labels every attack \textsc{safe}. The only correct, uniform distinguishing property is \emph{termination of the scan cycle}. The trustworthy label came from reading the mutation, not from assuming what should be verified. We generalize this discipline into a ground-truth methodology in which every verdict is established by one of three explicit, recorded methods -- by construction, fault injection, or audited cross-tool consensus -- never by an unexamined guess.

\medskip\noindent\textbf{Contributions}
\begin{itemize}
\item \textbf{A benchmark suite for IEC~61131-3 formal verification} (\S\ref{sec:suite}): 50 benchmarks / 83 variants across textual \gls{ld}, graphical \gls{ld}, and \gls{st}, in ten industrial fields, each with a formal property, a machine-checkable expected verdict, and a witness. It is, to our knowledge, the first such suite to cover graphical \gls{ld}.
\item \textbf{A ground-truth methodology} (\S\ref{sec:groundtruth}) that triangulates every verdict across construction, fault injection, and audited consensus, together with the concrete lesson -- from the non-terminating-loop episode -- that trustworthy labels come from inspecting mutations, not from assuming the intended property.
\item \textbf{A reusable, SV-COMP--aligned format and harness} (\S\ref{sec:principles}, \S\ref{sec:groundtruth}): a schema-validated task layout and a one-command runner that rechecks every verdict, letting new tools plug in via adapters and enabling a \gls{plc}/\gls{ics} competition category.
\item \textbf{Cross-tool baseline results} (\S\ref{sec:eval}) on the full suite: all 25 graphical benchmarks are verified by \esbmcplc{} (43/45 accepted variants match ground truth), and -- on the 21 finite-state benchmarks it can ingest -- an independent model checker (nuXmv) agrees on all 24 interlock variants and resolves two stateful benchmarks that \esbmcplc{} leaves unknown, an independent confirmation of the ground truth and a demonstration that the suite discriminates between tools. The evaluation also quantifies the input-format fragmentation that today prevents any single tool from running on all programs.
\end{itemize}

\medskip\noindent
The complete corpus, property files, witnesses, schema, validator, and recheck harness are archived under a citable DOI at the tool commit used for the baseline (\S\ref{sec:eval}).

\section{Related Work}
\label{sec:related}

\subsection{Formal Verification of IEC~61131-3}
A growing set of tools verifies \gls{plc} logic. The \esbmcplc{}~\citep{dantas_esbmc-plc_2026-3,dantas_esbmc-graphplc_2026,dantas_esbmc-plc_2026-1} provides an \gls{smt}-based bounded-model-checking and $k$-induction frontend for \gls{ld} and \gls{st}; PLCverif~\citep{tournier_plcverif_2022,fink_verifying_2024} at \gls{cern} targets \gls{st} and drives several back-end checkers; and general model checkers such as nuXmv~\citep{cavada_nuxmv_2014} and deductive tools such as Why3~\citep{why3} have been applied through translation. These are the tools our suite is built to serve. Their evaluations, however, use private or mutually incomparable program sets with ad hoc properties, so cross-tool claims cannot be checked -- precisely the gap that a shared, property-bearing suite closes. Our work is complementary rather than competing: we provide the common target and, in \S\ref{sec:eval}, run several of these tools on it.

\subsection{Benchmark Suites and Verification Competitions}
In software verification for C, SV-COMP~\citep{beyer_state_2024} and Test-Comp~\citep{testcomp} show how a standard suite with explicit expected verdicts, machine-readable task definitions, and validated witnesses can turn isolated tools into a comparable, cumulatively improving community. We adopt that task model and adapt it to \gls{plc} verification. Our contribution beyond that transfer is threefold, and none of it arises in SV-COMP's single-language, single-semantics C setting: coverage of \emph{graphical} \gls{ld}; a \emph{termination} property class discharged by a scan-cycle watchdog rather than an assertion; and a \emph{per-tool-verdict} stance for benchmarks whose verdict depends on a tool's scan-cycle timing semantics. For IEC~61131-3 specifically, the closest prior effort is the \plcopen{} benchmark suite of \citet{ukegbu_benchmarks_2023}, which pairs programs with properties but covers \gls{st} only. Our suite differs in two respects: it targets graphical \gls{ld} (the encoding real programs are exchanged in and the one no prior suite addresses), and it makes ground-truth provenance an explicit, recorded part of every task (\S\ref{sec:groundtruth}) rather than an implicit assumption.

\subsection{\Gls{plc} Program Corpora and \gls{ics} Security Datasets}
Several program collections exist, but are not verification tasks. The \plcld{} tank-control dataset~\citep{iacobelli_detection_2024} and the SWaT corpus~\citep{rinieri_plc_defuser_2024} (which we ingest via the \texttt{PLC\_Defuser} distribution) supply ``legitimate''/``malicious'' program pairs but no formal properties or expected verdicts; the malicious variants realize the Ladder Logic Bomb threat model of \citet{govil_ladder_2017}. The \gls{ppu} clone-detection corpus~\citep{rosiak_iec_2022} provides graphical programs for a different task entirely (clone detection) and likewise ships no properties. We reuse programs from all three, but our contribution is orthogonal to theirs: we attach a formal property and a triangulated expected verdict to each program, thereby converting a program corpus into a verification suite. This step is where the non-terminating-loop finding of \S\ref{sec:groundtruth} arose -- the ``malicious'' label alone does not tell a verifier \emph{which} property is violated.

\subsection{Formal Semantics of IEC~61131-3}
Our properties are stated over a per-scan execution model whose semantics rests on prior formalization work: the K-ST executable semantics of \citet{wang_k-st_2023}, the \gls{ld} translation rules of \citet{ebnenasir_formalizing_2023}, and the MATIEC compiler~\citep{de_sousa_matiec_2014}, which is used by several toolchains. The 15 feature micro-benchmarks in our suite are drawn from an independent executable \gls{ld}-semantics effort, which is why we treat their verdicts as tool-independent (\S\ref{sec:groundtruth}). That line of work targets the \emph{meaning} of the languages; ours targets a shared, verdict-bearing corpus for evaluating the tools that check them.

\section{Design Principles}
\label{sec:principles}

The community has no standard benchmark suite for formal verification of IEC~61131-3 \gls{ld} programs. Existing resources are lacking in one of two ways: program corpora (e.g.\ the \plcld{} dataset, the \gls{ppu} clone-detection set) ship code without formal properties or expected verdicts, and are therefore unusable as verification tasks; and the one property-bearing suite we are aware of 
\citet{ukegbu_benchmarks_2023} covers \gls{st} only, omitting the graphical \gls{ld} encoding that dominates the installed base. We therefore designed the suite from first principles, distilling six requirements from established benchmarking practice -- chiefly the SV-COMP task model of an explicit \emph{expected verdict}, a machine-readable task definition, and a witness for every violation -- and adapting them to the specifics of \gls{plc} verification. Each principle is stated below with the design decision it drove; \S\ref{sec:suite}--\ref{sec:groundtruth} reports how each is realized, and \S\ref{sec:threats} the residual risk.

\begin{enumerate}[label=\textbf{P\arabic*.
}, leftmargin=*, align=left]
    \item \textbf{Ground truth is fundamental, machine-checkable, and triangulated.} A benchmark's entire value rests on the trustworthiness of its expected verdicts: a single mislabelled task silently rewards a wrong answer and penalizes a right one. We treat the verdict, not the program, as the primary artifact. Every verdict is (i)~carried in a schema-validated field rather than a comment, so it is machine-checkable and cannot drift from the program; and (ii)~established by one of three explicit, recorded methods -- by construction, fault injection, or audited cross-tool consensus (\S\ref{sec:groundtruth}) -- so that no label rests on a single unexamined judgement. This principle overrides the others: where realism or coverage would have forced an unverifiable verdict, we excluded the task from consideration.

    \item \textbf{A verification task is a program \emph{and} a formal property.}
    What distinguishes a verification suite from a program corpus is that each problem formally states \emph{what is to be proved}. Every benchmark, therefore, pairs its program with one or more properties drawn from a small, closed vocabulary (invariant, mutual-exclusion, reachability, termination), expressed over program variables under the per-scan execution model. Fixing a closed vocabulary keeps tasks comparable across tools and prevents the property language from silently becoming a second, unspecified dimension of difficulty.

    \item \textbf{Diversity across language, domain, and difficulty.} A uniform suite measures one narrow capability. We require a spread along three axes. \emph{Language:} all three practically used IEC~61131-3 forms (textual and graphical \gls{ld} and \gls{st}) with deliberate emphasis on graphical \gls{ld}, the encoding of real programs is exchanged in, and the one with no prior suite targets. \emph{Domain:} ten industrial sectors, so that results are not an artifact of a single process type. \emph{Difficulty:} a gradient from single-construct micro-benchmarks to multi-\gls{plc} controllers, allowing the suite to separate tools rather than saturate at a single level.

    \item \textbf{Discrimination over triviality.}
    A suite on which every tool scores perfectly is uninformative. We therefore balance the expected verdicts (roughly half \textsc{safe}, half \textsc{violation}), so that a strategy of always answering ``safe'' fails half the tasks, and we retain a hard tier whose members' current tools are not guaranteed to solve. Balanced verdicts also force tools to demonstrate both proof and refutation, and expose the \emph{unknown} outcomes that a safe-only suite would hide.

    \item \textbf{Realistic provenance, with the authored share disclosed.}
    Realism and control are in tension: real programs are representative but lack properties or clean fault labels, while authored programs are controllable but risk tuning the suite to the authors' own tools. We address this by preferring third-party and prior programs wherever they exist, authoring only to fill genuine gaps (domains and constructs with no public program), and disclosing the exact split (\S\ref{sec:suite}) so that a reader can weigh the two. Authored programs follow standard control idioms -- interlocks, guarded actuation -- rather than any tool-specific pattern.

    \item \textbf{A reusable, tool-agnostic format and one-command recheck.}
    A suite becomes a standard only if others can run it. We adopt a stable, documented, schema-validated layout aligned with the SV-COMP task model (task descriptor, property file, per-violation witness), so that a new tool is added through an adapter rather than by reformatting the corpus. So the suite can seed an SV-COMP \gls{plc}/\gls{ics} category directly. Reproducibility is built in: a single harness re-runs every task and checks the result against its recorded verdict, and where a tool cannot ingest a program, we record that explicitly rather than dropping the task, keeping coverage honest.

\end{enumerate}

\section{The Benchmark Suite}
\label{sec:suite}

\begin{figure}[htbp]
\centering
\begin{tikzpicture}[
  font=\footnotesize, >={Stealth[length=2.2mm]},
  task/.style={draw, rounded corners=2pt, fill=black!5,  align=left,   inner sep=4pt, text width=75mm},
  val/.style ={draw, rounded corners=2pt, fill=black!5,  align=center, inner sep=3pt, text width=16mm},
  hub/.style ={draw, rounded corners=2pt, fill=black!12, align=center, inner sep=3pt, minimum width=50mm, minimum height=6mm},
  tool/.style={draw, rounded corners=2pt, fill=blue!8,   align=center, inner sep=3pt, text width=31mm, minimum height=11mm},
  result/.style ={draw, rounded corners=2pt, fill=green!10, align=center, inner sep=4pt, text width=100mm},
]
\node[tool] (t1) {\textbf{\esbmcplc{}}\\[1pt]\emph{native} \plcopen{}~XML};
\node[tool, right=6mm of t1] (t2) {\textbf{nuXmv}\\[1pt]translate $\to$ \gls{smv}};
\node[tool, right=6mm of t2] (t3) {\textbf{PLCverif}\\[1pt]translate $\to$ \gls{st}};
\node[hub, above=9mm of t2] (hub) {runner / one-command re-check harness};
\node[task, above=8mm of hub, xshift=-0mm, anchor=south] (task) {%
  \textbf{Benchmark task} (SV-COMP-compatible)
  program:\\
    - \plcopen{}~XML \emph{or} \gls{st} \\
    - witness property: YAML $\langle$invariant, mutual\_excl., reachability, termination$\rangle$\\
    - machine-checkable \textbf{expected verdict}};
\node[val, right=4mm of task, anchor=west] (val) {JSON-Schema\\validator};
\node[result, below=8mm of t2] (out) {%
  verdict $\in \{\textsc{safe},\ \textsc{violation},\ \textsc{unknown}\}$\ \ vs.\ recorded ground truth\\
  $\Rightarrow$\ \textbf{agreement} (tool-neutral verdict) \ and \ \\\textbf{discrimination}
  (one tool resolves another's \textsc{unknown})};
\draw[->] (task.east) -- (val.west);
\draw[->] (task) -- (hub);
\draw[->] (hub) -- (t1.north);
\draw[->,dashed] (hub) -- (t2.north);
\draw[->,dashed] (hub) -- (t3.north);
\draw[->] (t1.south) -- (t1.south |- out.north);
\draw[->] (t2.south) -- (out.north);
\draw[->] (t3.south) -- (t3.south |- out.north);
\end{tikzpicture}
\caption{A benchmark task and the cross-tool runner. Each task pairs a program (\plcopen{}~XML or \gls{st}) with a formal property, a machine-checkable expected verdict, and a witness, all of which are validated against a JSON Schema. The runner dispatches each task to a tool through an adapter: \esbmcplc{} ingests \plcopen{}~XML \emph{natively} (solid), whereas nuXmv and PLCverif require translation to \gls{smv} and \gls{st} (dashed) -- the format fragmentation the suite makes measurable. Each tool's verdict is compared to the recorded ground truth, yielding the cross-tool agreement and discrimination results of \S\ref{sec:eval}.} 
\label{fig:pipeline}
\end{figure}
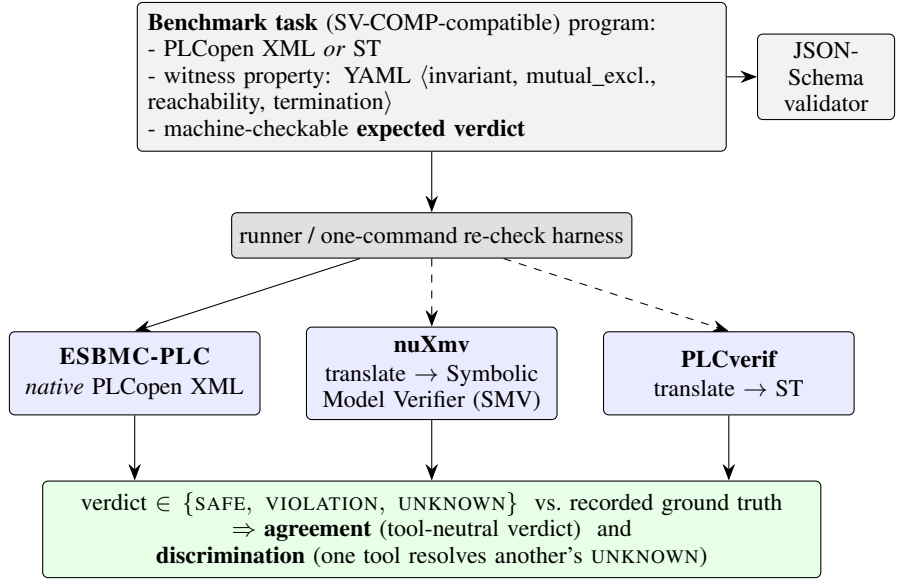

The suite comprises \textbf{50 benchmarks} over \textbf{83 program variants}, each an IEC~61131-3 program in standard \plcopen{}~XML \gls{ld} or \gls{st}, paired with a machine-checkable property file and an expected verdict (Fig.~\ref{fig:pipeline}). A \emph{benchmark} is a verification task; a \emph{variant} is one program file under that task (30~benchmarks bundle a safe program with one or more faulted counterparts, \S\ref{sec:groundtruth}). Every benchmark is validated against the suite's JSON Schema. It is packaged in an SV-COMP--compatible layout (task descriptor, property file, witness fields), so the artifact can seed a future SV-COMP \gls{plc}/\gls{ics} category without rework.

\subsection{Composition}
Table~\ref{tab:composition} provides a breakdown by IEC~61131-3 language and industrial domain. The language split (15~textual~\gls{ld}, 25~graphical~\gls{ld}, 10~\gls{st}) deliberately favors \emph{graphical} \gls{ld}: it is the encoding in which real, third-party programs are actually exchanged, and no prior formal-verification suite targets it (the closest, the \plcopen{} suite of~\citet{ukegbu_benchmarks_2023}, covers 40~\gls{st} programs only). The suite spans \textbf{ten industrial fields}; water treatment is the largest (\S\ref{sec:threats}) because the two richest public \gls{ics} corpora -- SWaT and the \plcld{} tank-control dataset -- are both in the water domain.

\begin{table}[htbp]
  \centering\small
  \caption{Suite composition by language and by domain (50 benchmarks).}
  \label{tab:composition}
  \begin{tabular}{@{}llr@{\hspace{2em}}lr@{}}
    \toprule
    \multicolumn{2}{@{}l}{\textbf{Language}} & \textbf{\#} &
    \textbf{Domain} & \textbf{\#}\\
    \midrule
    \gls{ld} & (textual)   & 15 & water treatment    & 16\\
    \gls{ld} & (graphical) & 25 & manufacturing      & 6\\
    \gls{st}&             & 10 & motor control      & 5\\
    \cmidrule(r){1-3}
    \multicolumn{2}{@{}l}{\textbf{Total}} & \textbf{50} & packaging & 5\\
                   &             &    & traffic            & 4\\
    \multicolumn{2}{@{}l}{\textbf{Difficulty}} &  & building automation & 3\\
    easy           &             & 26 & elevator           & 3\\
    medium         &             & 18 & HVAC               & 3\\
    hard           &             &  6 & power substation   & 3\\
                   &             &    & chemical batch     & 2\\
    \bottomrule
  \end{tabular}
\end{table}

\subsection{Provenance}
To balance realism against controlled coverage, the suite draws from four external or prior sources and supplements them with authored programs (Table~\ref{tab:provenance}). Nineteen benchmarks come from third-party or prior corpora: the \plcld{} tank-control dataset of \citet{iacobelli_detection_2024}, the SWaT testbed via the \texttt{PLC\_Defuser} dataset, five curated \esbmcplc{} security scenarios, and one program from the \gls{ppu} clone-detection corpus. The remaining 31 are authored: 15 are feature micro-benchmarks that systematically exercise individual \gls{ld} constructs (contacts, coils, seal-in, \texttt{TON}/\texttt{TOF}/\texttt{TP} timers, \texttt{CTU}/\texttt{CTD} counters, edge detectors, latches), and 16 are domain control programs written to give each of the ten domains a genuine, non-analogical anchor -- particularly the domains for which no public program was available (chemical batch, power substation).

\begin{table}[htbp]
  \centering\small
  \caption{Provenance of the 50 benchmarks.}
  \label{tab:provenance}
  \begin{tabular}{@{}lr l@{}}
    \toprule
    \textbf{Source} & \textbf{\#} & \textbf{Nature}\\
    \midrule
    \plcld{} dataset (Iacobelli et al.) & 8  & graphical \gls{ld}, real third-party\\
    SWaT / \texttt{PLC\_Defuser}               & 6  & multi-\gls{plc} \gls{st}, real testbed\\
    \esbmcplc{} security scenarios      & 4  & curated \gls{st}\\
    \gls{ppu} clone-detection corpus        & 1  & graphical \gls{ld}\\
    Authored: feature micro-benchmarks& 15 & one \gls{ld} construct each\\
    Authored: domain control programs & 16 & interlocks + 5 syntax twins\\
    \midrule
    \textbf{Total}                    & \textbf{50} & \\
    \bottomrule
  \end{tabular}
\end{table}

\subsection{Properties and coverage}
Each benchmark carries one or more formal properties in a YAML schema with a closed vocabulary of kinds: \emph{invariant} (29 properties), \emph{termination} (14), \emph{mutual\_exclusion} (6), and \emph{reachability} (3, used for counterexample/trigger synthesis). Properties are expressed over program variables under the per-scan execution model. Apart from safety invariants and interlocks, the suite covers a property class absent from existing \gls{plc} benchmarks: \emph{termination} of the scan cycle, which is the safety-relevant manifestation of the non-terminating-loop logic bombs described in \S\ref{sec:groundtruth}.

\subsection{Syntax-coverage pairs}
Five benchmarks are graphical-\gls{ld} renderings of a textual-\gls{ld} or \gls{st} original with identical logic (recorded via a \texttt{syntax\_pair\_of} field). These are not padding: they let the suite act as a \emph{differential} test of whether a tool that accepts one \plcopen{} encoding produces the same verdict on the other, a discrepancy we consider a defect. They are counted once for each language slice and explicitly flagged so that a reader can include or exclude them.

\section{Ground-Truth Methodology}
\label{sec:groundtruth}

A benchmark suite is only as useful as its expected verdicts: a single mislabelled task silently rewards wrong answers. We therefore establish each verdict using one of three methods and record the method used for each variant in the benchmark descriptor. Across the 83 variants, the expected verdicts are 49~\textsc{safe} and 34~\textsc{violation}; the three methods account for 36, 29, and 18 variants respectively (Table~\ref{tab:groundtruth}).

\begin{table}[htbp]
  \centering\small
  \caption{Ground-truth method by variant (83 variants).}
  \label{tab:groundtruth}
  \begin{tabular}{@{}lr l@{}}
    \toprule
    \textbf{Method} & \textbf{\#} & \textbf{Basis of the verdict}\\
    \midrule
    Expert / by-construction     & 36 & property holds (or fails) by the program's design\\
    Fault injection              & 29 & mutant of a safe base with a documented defect\\
    Cross-tool consensus         & 18 & $\geq$2 tools agree, plus manual audit\\
    \bottomrule
  \end{tabular}
\end{table}

The feature micro-benchmarks and the authored domain interlocks are constructed so that their properties are determined by inspection: a series interlock makes two outputs mutually exclusive; a not-full contact guards a fill-valve rung. For these, we record the verdict as \textsc{expert} and give the one-line justification in the property file.

For 29 benchmarks, the \textsc{violation} variant is a \emph{mutant} of a known-safe base, with the seeded defect and a triggering witness documented in the descriptor. This is the most trustworthy source of \textsc{violation} labels because the defect is known by construction. Crucially, both public corpora we ingest already supply such pairs: every ``malicious'' program in the \plcld{} dataset and in SWaT is a fault-injected sibling of a ``legitimate'' one.

Inspecting these mutants surfaced a uniform, and initially non-obvious, fact: the injected defect in \emph{every} malicious variant of both corpora is a \emph{non-terminating loop} -- a \texttt{WHILE} whose induction variable never reaches its bound (e.g.\ \texttt{i := i*i} from zero, or a counter that is never advanced), gated on a specific sensor value. The controllers' functional logic (valve interlocks, pump hysteresis) is left untouched. A na\"ive labelling that asserted a valve-safety property would therefore have marked these mutants \textsc{safe}, because that property still holds -- the label would have been wrong. The correct, uniform distinguishing property is \emph{termination of the scan cycle}: the safe program completes every scan, the mutant hangs once the trigger input occurs. We encode this as the \texttt{termination} property kind, which a bounded model checker discharges via a scan-cycle watchdog. This episode is the strongest argument for the method: the trustworthy label came only from reading the injected code, not from assuming what the ``obvious'' safety property should be.

For the \textsc{safe} variants of the fault-injection pairs, where no by-construction argument fixes the verdict, we require agreement between independent tools followed by manual audit of any disagreement, and record the confirming tools in the descriptor. Disagreements are not discarded: they are adjudicated by hand and reported, since a program on which mature tools disagree is among the most informative in the suite.

Verdicts are re-checkable end to end: \texttt{run\_all.sh} runs every variant through the verifier -- \texttt{-{}-k-induction} for expected-\textsc{safe} tasks, \texttt{-{}-incremental-bmc} for expected-\textsc{violation} tasks, with the scan watchdog enabled -- and compares the result to the recorded verdict, emitting a per-variant pass/fail table. The full suite, property files, witnesses, and this harness are archived with a citable DOI.

\section{Evaluation}
\label{sec:eval}

We determine reference verdicts on the suite with \esbmcplc{}, built from source with the \gls{ld} frontend enabled (\texttt{-DENABLE\_LD\_FRONTEND=On}); expected-\textsc{safe} tasks are run under $k$-induction and expected-\textsc{violation} tasks under incremental BMC, both with the scan-cycle watchdog enabled. The harness attempts all 50 benchmarks and records the outcome of every variant against its recorded verdict.

\subsection{Results}
Of the 83 variants, the 45 that the frontend accepts -- all the 25 graphical \plcopen{}~XML benchmarks -- run, and \textbf{43 match the recorded verdict with zero disagreements}; the two non-matches are \textsc{unknown} outcomes ($k$-induction proof-strength results on safe edge-detector programs, reported as \textsc{unknown} rather than folded into \textsc{safe}, per \S\ref{sec:threats}). This confirms the recorded ground truth across every authored domain interlock (safe variant proved, faulted variant refuted with a counterexample) and every non-termination tank controller (the injected logic bomb caught by the scan watchdog). It is clear, tool-checked evidence that the property files, the seeded defects, and the witnesses are consistent. Reaching this required aligning the property syntax to v8.4's C-style boolean operators and running an adapter that maps the suite schema to the tool's property format -- the adapter is released with the suite.

\subsection{A Format-and-Semantics Finding}
The remaining 38 variants -- the textual-Ladder and Structured-Text benchmarks -- did not run, and the reason is more than skin-deep. First, format: v8.4's frontend accepts \emph{only} \plcopen{}~XML, so a program in the textual \gls{ld} form or a stand-alone \gls{st} module is rejected outright. Second, and more interesting, \emph{semantics}: when we transcribed the textual timer and counter benchmarks into \plcopen{}~XML, their recorded verdicts -- established under an independent executable \gls{ld} semantics -- did \emph{not} transfer, because \esbmcplc{}'s scan-cycle models of \texttt{TON}/ \texttt{CTU} and of \texttt{set}/\texttt{reset} latches differ from that reference. Porting these benchmarks, therefore, requires re-establishing the ground truth under each tool's timing model, rather than a mechanical re-serialization. Both are concrete instances of the fragmentation this suite is designed to expose -- one tool's frontend admits a single serialization, and even the meaning of a standard timer is not agreed -- and both motivate the suite carrying every program in vendor-neutral \plcopen{}~XML \emph{with per-tool verdicts where the semantics diverge}, which is the principal item of remaining work.

\subsection{Cross-Tool Confirmation}
A verdict established partly by the tool under evaluation is only as convincing as its independence. We therefore re-verified the twelve combinational interlock benchmarks with \textbf{nuXmv~2.2.0}~\citep{cavada_nuxmv_2014}, whose \gls{bdd}/\gls{ic3} decision procedure is unrelated to \gls{esbmc}'s \gls{smt}-based bounded model checking. Because nuXmv accepts only its own \gls{smv} language, each program was translated from \plcopen{}~XML into an \gls{smv} module (inputs as free Boolean variables, coils as defined functions, the safety property as an \texttt{INVARSPEC}) -- a translation that is itself an instance of the fragmentation this suite documents. The \gls{smv} models are generated systematically from the same rung structure as the \plcopen{}~XML and are released with the suite; their agreement with \esbmcplc{} on the \emph{native} XML is itself a cross-check on the translation, since a faulty transcription would in general perturb the verdict rather than preserve it across two unrelated engines. 

On all \textbf{24 variants nuXmv agrees with \esbmcplc{} and with the recorded ground truth}: every safe variant is proved, and every faulted variant is refuted. We additionally modeled three \emph{stateful} benchmarks -- a seal-in latch and two edge detectors -- as \gls{smv} transition systems that mirror the scan cycle (a persistent state variable per retained output, updated by \texttt{next()}). nuXmv proves all three safe and, tellingly, \textbf{resolves the two edge-detector benchmarks that \esbmcplc{}'s $k$-induction reports as \textsc{unknown}} (Table~\ref{tab:crosstool}). This is the clearest possible demonstration that the suite serves its purpose: two mature tools with independent decision procedures reaching identical verdicts is strong evidence that the ground truth is tool-neutral rather than an artefact of the tool that helped establish it; and where the two tools \emph{differ} -- one proving what the other leaves open -- the suite has discriminated between them, which is exactly what a benchmark is for.

\begin{table}[htbp]
  \centering\small
  \caption{Cross-tool verdicts. \esbmcplc{} (\gls{smt}-\gls{bmc}/$k$-induction, native \plcopen{}~XML) and
           nuXmv~2.2.0 (\gls{bdd}/\gls{ic3}, on translated \gls{smv}) agree with the ground truth wherever both
           conclude; nuXmv proves two edge-detector benchmarks that \esbmcplc{}'s $k$-induction
           leaves open.}
  \label{tab:crosstool}
  \begin{tabular}{@{}lccc@{}}
    \toprule
    & \textbf{\esbmcplc{}} & \textbf{nuXmv} & \textbf{Ground truth}\\
    \midrule
    interlocks: safe (12)        & 12 proved      & 12 proved       & \textsc{safe}\\
    interlocks: faulted (12)     & 12 refuted     & 12 refuted      & \textsc{violation}\\
    stateful: seal-in (1)        & proved         & proved          & \textsc{safe}\\
    stateful: edge detectors (2) & \emph{unknown} & \textbf{proved} & \textsc{safe}\\
    \bottomrule
  \end{tabular}
\end{table}

\subsection{Scope of These Numbers}
The cross-tool comparison covers the fifteen graphical benchmarks whose logic is finite-state (combinational interlocks and the seal-in/edge-detector automata). The remaining ten graphical benchmarks are the non-terminating-loop tank controllers, whose defect is a liveness property over unbounded arithmetic that we do not encode in a finite-state \gls{smv} model; they, along with the textual-\gls{ld} and \gls{st} benchmarks, remain single-tool. PLCverif, which consumes \gls{st} rather than \plcopen{}~XML, likewise requires translation. A complete multi-tool matrix across all three languages -- and the \plcopen{}~XML normalization that would let each tool ingest the whole suite -- is the principal item of remaining work, and is exactly the fragmentation the suite is built to make measurable.

\subsection{Validation Coverage}
We record in the schema (\texttt{validation\_status}) whether each benchmark's verdict is \emph{tool-confirmed} or \emph{candidate} -- established by construction or fault injection but not yet independently checked by a model checker. Of the 50 benchmarks, \textbf{31 are validated}: the 25 graphical benchmarks (\esbmcplc{}, 15 also confirmed by nuXmv) and the six semantics-independent textual-\gls{ld} benchmarks (Boolean and edge/latch logic, confirmed by nuXmv). The remaining \textbf{19 are candidates}: the nine timer/counter/latch feature benchmarks, whose verdicts are \emph{semantics-dependent} -- our porting experiments show a standard timer's verdict can change across tools' scan-cycle timing models -- and the ten Structured-Text programs, whose non-termination and multi-\gls{pou} structure we do not yet encode for a second checker. 

The empirical claims above concern the validated set; the candidate verdicts are flagged as such so that no result depends on an unconfirmed label, and their tool confirmation follows the porting protocol released with the suite. The validated verdicts span all ten domains and both provenance classes (22 of 31 authored, 9 of 19 third-party); the candidate set is not concentrated in any industrial domain but in two program \emph{kinds} -- the timer/counter/latch feature benchmarks and the Structured-Text programs -- and each benchmark's status is recorded per task so that a reader can reweight results by domain or provenance from the released data. Marking semantics-dependent verdicts \emph{candidate} rather than forcing them through one tool's timer model is itself a consequence of the paper's thesis: where tools disagree on meaning, a single ``ground truth'' is the wrong abstraction.

\section{Threats to Validity}
\label{sec:threats}

We discuss threats along the standard construct, internal, external, and reproducibility axes, and state the mitigations we adopted and the residual risk.

\subsection{Construct Validity: Do the Tasks Measure Verification Capability?}
The properties are written by us, so the suite risks measuring ``can the tool prove \emph{the properties we chose}'' rather than program correctness. We reduce this through deriving each property from the program's own control logic (interlocks, guarded actuation, scan termination) rather than inventing abstract assertions, and by drawing from a closed, documented vocabulary (\S\ref{sec:suite}). Most benchmarks carry a single property; the suite is not intended to certify full functional correctness, only to provide decidable, well-motivated safety obligations. A second construct threat is \emph{fault-model coverage}. The 34 \textsc{violation} variants comprise 14 scan non-terminations, 14 safety-invariant violations, and 6 interlock (mutual-exclusion) violations. Importantly, the two public logic-bomb corpora we ingest (\plcld{} and SWaT) inject \emph{only} non-terminating loops; every other defect class in the suite (interlock bypass, actuator manipulation, overflow-guard evasion) is represented through \emph{authored or curated} mutants. Real-world violation examples in the suite are thus narrower in defect type than the authored ones, a limitation inherited from the available public corpora.

\subsection{Internal Validity: Are the Expected Verdicts Correct?}
A mislabeled task is the most damaging defect a suite can have. Our three-method triangulation (\S\ref{sec:groundtruth}) targets this directly: fault-injection labels are known by construction, expert labels are decidable by inspection, and consensus labels require independent agreement plus manual audit. The non-terminating-loop episode (\S\ref{sec:groundtruth}) is evidence that the audit process catches mislabelling that a plausible-but-wrong property would have introduced. Two residual risks remain. First, the expert and fault-injection labels rely on our reading of the programs; we mitigate by publishing every justification and witness so they are independently checkable, by the end-to-end recheck below, and by explicitly marking as \emph{candidate} the 19 benchmarks whose verdicts are not yet tool-confirmed (\S\ref{sec:eval}) -- so the 31 validated verdicts, on which the empirical claims rest, are separated in the schema from those established by authorial judgement alone. Second, a \emph{consensus} \textsc{safe} label and the tool under evaluation are not fully independent: \esbmcplc{} participates in both establishing consensus and serving as the baseline. We limit this by requiring at least one tool-independent basis (either by construction or a second, architecturally different checker) for every \textsc{safe} label, and by noting that only 18 of 83 variants rest on consensus at all. We further distinguish \emph{proof strength} from \emph{truth}: a $k$-induction run may return \emph{unknown} on a genuinely safe program; these outcomes are reported as \textsc{unknown}, not silently folded into \textsc{safe}.

\subsection{External Validity: Do Results Generalize?}
Two facts bound generalization. First, 31 of 50 benchmarks are authored, which risks a suite tuned to constructs our own frontend handles well; we counter this by including 19 third-party/prior programs, by sourcing the authored feature benchmarks from an independent executable semantics effort, and by writing the authored domain programs to standard interlock patterns rather than to our tool. Second, the domain distribution is skewed: water treatment accounts for 16 of 50 benchmarks, because the two richest public \gls{ics} corpora are both water-domain. We report all results per domain so that a reader can reweight them, and we ensured that each of the ten domains has at least two genuine anchors, so none is represented only by analogy. Program scale is a further external threat: many benchmarks are small, and heavy scalability stress rests on the multi-\gls{plc} SWaT programs and the counter-fuse scalability family alone. Finally, five benchmarks are syntax-coverage twins; they are flagged and counted once per language slice, so their shared logic cannot inflate coverage claims.

\subsection{Reproducibility}
All verdicts are re-checkable with a single command (\texttt{run\_all.sh},
\S\ref{sec:groundtruth}), which selects the proof mode based on the expected verdict and compares the tool's result with the recorded label. Two practical caveats apply to cross-tool baselining: the baseline checkers consume different input formats, so no single tool runs on all 50 programs, and we report per-tool coverage explicitly rather than dropping unsupported rows -- the format fragmentation itself is a finding that motivates a common suite. The complete corpus, property files, witnesses, schema, and harness are archived under a citable DOI at the exact tool commit used for the baseline.

\section{Conclusion}
\label{sec:conclusion}

Formal verification of IEC~61131-3 programs has active tools but lacks a common yardstick: without a shared, property-bearing benchmark suite, progress cannot be measured, and techniques cannot be compared. We have presented such a suite -- 50 benchmarks over 83 program variants, spanning textual and graphical \gls{ld} and \gls{st} across ten industrial fields, each carrying a formal property, a machine-checkable expected verdict, and, for every violation, a triggering witness. To our knowledge, it is the first formal-verification benchmark suite to cover graphical \gls{ld}, the encoding used to exchange real programs. The corpus is packaged in an SV-COMP--compatible layout with a schema validator and a one-command recheck harness, so that a new tool can be added via an adapter and the artifact can directly seed a \gls{plc}/\gls{ics} category in SV-COMP.

Our central lesson concerns ground truth. For programs of any realism, the correct verdict is not self-evident, and an authoritative-looking label can be quietly wrong: the ``malicious'' variants of two public corpora would have been mislabelled \textsc{safe} by the obvious valve-interlock property, because the injected defect is a non-terminating loop that leaves the valve logic intact. The trustworthy label came only from reading the mutation. We therefore make ground-truth provenance a first-class, recorded property of every task, established by construction, fault injection, or audited cross-tool consensus -- a discipline we believe any verification benchmark, in any domain, should adopt.

The suite is designed to grow. Two commitments follow directly from this release. First, we will tool-confirm the nineteen candidate verdicts under each tool's own semantics, using the porting protocol shipped with the suite -- accepting, per our per-tool-verdict stance, that a timer-dependent verdict may legitimately differ between tools and recording both. Second, once the corpus is normalized to a single \plcopen{}~XML form that every frontend accepts, we will run a full multi-tool matrix -- PLCverif and nuXmv alongside \esbmcplc{} -- turning the partial cross-tool comparison of \S\ref{sec:eval} into a complete one. Beyond these, the natural following steps are to broaden the fault model beyond the non-termination bombs that dominate the available public corpora, add programs exported from proprietary vendor toolchains to widen manufacturer coverage, and enlarge the corpus to the scale needed for a standing competition category. By releasing the benchmarks, property files, witnesses, schema, and harness under a citable DOI, we intend the suite to serve as a shared, extensible foundation on which the \gls{plc} formal-verification community can measure and compare its progress.

\section*{Acknowledgements}
The authors thank the ESBMC and PLCverif developers for their tools and documentation.

\section*{Declarations}

\textbf{Funding.} This work was partially funded by the Engineering and Physical Sciences Research Council (EPSRC) grants EP/T026995/1, EP/V000497/1, and EP/X037290/1, and by the Soteria project, awarded by UK Research and Innovation under the Digital Security by Design (DSbD) Programme.\\[2pt]

\textbf{Data availability.} The complete benchmark suite -- programs, property files, witnesses, schema, validator, and re-check harness -- is openly available at \url{https://github.com/pierredantas/esbmc-plc-benchmark-suite}.\\[2pt]

\bibliographystyle{unsrtnat}   
\bibliography{refs}

\end{document}